\documentclass[journal]{IEEEtran}

\usepackage{enumerate}
\usepackage{cite}
\usepackage[pdftex]{graphicx}
\usepackage{amsmath}
\usepackage{amssymb}
\usepackage{array}
\usepackage{xcolor}

\usepackage{bm}
\usepackage{booktabs}
\usepackage[caption=false]{subfig}
\begin{document}

\title{Quantum Language Model with Entanglement Embedding for Question Answering}


\author{Yiwei~Chen, 
        Yu~Pan,~\IEEEmembership{Senior~Member,~IEEE,}
        Daoyi~Dong,~\IEEEmembership{Senior~Member,~IEEE}
\thanks{This work was supported by the National Natural Science Foundation of China
(No. 61703364) and the Australian Research Council's Discovery Projects funding scheme under Project
DP190101566.}
\thanks{Y. Chen is with the Institute of Cyber-Systems and Control, Zhejiang University, Hangzhou, 310027, China. (email: ewell@zju.edu.cn).}
\thanks{Y. Pan is with the Institute of Cyber-Systems and Control, College of Control Science and Engineering, Zhejiang University, Hangzhou, 310027, China. (email: ypan@zju.edu.cn).}
\thanks{D. Dong is with the School of Engineering and Information
Technology, University of New South Wales, Canberra, ACT 2600,
Australia. (email: daoyidong@gmail.com).}

}

\maketitle

\begin{abstract}
Quantum Language Models (QLMs) in which words are modelled as a quantum superposition of sememes have demonstrated a high level of model transparency and good post-hoc interpretability. Nevertheless, in the current literature word sequences are basically modelled as a classical mixture of word states, which cannot fully exploit the potential of a quantum probabilistic description. A quantum-inspired neural network module is yet to be developed to explicitly capture the non-classical correlations within the word sequences. We propose a neural network model with a novel Entanglement Embedding (EE) module, whose function is to transform the word sequence into an entangled pure state representation. Strong quantum entanglement, which is the central concept of quantum information and an indication of parallelized correlations among the words, is observed within the word sequences. The proposed QLM with EE (QLM-EE) is proposed to implement on classical computing devices with a quantum-inspired neural network structure, and numerical experiments show that QLM-EE achieves superior performance compared with the classical deep neural network models and other QLMs on Question Answering (QA) datasets. In addition, the post-hoc interpretability of the model can be improved by quantifying the degree of entanglement among the word states.
\end{abstract}

\begin{IEEEkeywords}
quantum language model, complex-valued neural network, interpretability, entanglement embedding.
\end{IEEEkeywords}

%
\IEEEpeerreviewmaketitle

\section{Introduction}
\IEEEPARstart{N}{eural} Network Language Model (NNLM) \cite{bengio2003neural} is widely used in Natural Language Processing (NLP) and information retrieval \cite{WQ20}. With the rapid development of deep learning models, NNLMs have achieved unparalleled success on a wide range of tasks \cite{mikolov2011extensions,mikolov2013efficient,sutskever2014sequence,he2016pairwise,Otter20Survey,WK20,DV20,wang2021emotion}. It becomes a common practice for the NNLMs to use word embedding \cite{mikolov2013distributed,mikolov2013efficient} to obtain the representations of words in a feature space. While NNLM has been very successful at knowledge representation and reasoning \cite{Otter20Survey}, its interpretability is often in question, making it inapplicable to critical areas such as the credit scoring system \cite{doshi2017towards}. Two important factors have been summarized in \cite{Lipton2018TheMO} for evaluating the interpretability of a machine learning model, namely, Transparency and Post-hoc Interpretability. The model transparency relates to the forward modelling process, while the post-hoc interpretability is the ability to unearth useful and explainable knowledge from a learned model.

Another emerging area is quantum information and quantum computation where quantum theory can be utilized to develop more powerful computers and more secure quantum communication systems than their classical counterparts \cite{nielsen2010quantum}. The interaction between quantum theory and machine learning has also been extensively explored in recent years. On one hand, many advanced machine learning algorithms have been applied to quantum control, quantum error correction and quantum experiment design \cite{dong2019learning,dunjko2018machine,chen2016quantum,chen2013fidelity}.
On the other hand, many novel quantum machine learning algorithms such as quantum neural networks and quantum reinforcement learning have been developed by taking advantage of the unique characteristics of quantum theory \cite{biamonte2017quantum,du2020learnability,havlivcek2019supervised,mcclean2018barren,nguyen2019benchmarking,li2020quantum,dunjko2016quantum, wei2021deep}.
Recently, Quantum Language Models (QLMs) inspired by quantum theory (especially quantum probability theory) have been proposed \cite{coecke2010mathematical,zeng2016quantum,sordoni2013modeling,sordoni2014learning,basile2017towards,zhang2018end,zhang2018quantum,li2019cnm} and demonstrated considerable performance improvement in model accuracy and interpretability on information retrieval and NLP tasks. QLM is a quantum heuristic Neural Network (NN) defined on a Hilbert space which models language units, e.g., words and phrases, as quantum states. By embedding the words as quantum states, QLM tries to provide a quantum probabilistic interpretation of the multiple meanings of words within the context of a sentence. Compared with the classical NNLM, the word states in QLM are defined on a Hilbert space which is different from the classical probability space. In addition, modelling the process of feature extraction as quantum measurement which collapses the superposed state to a definite meaning within the context of a sentence could increase the transparency of the model.

\begin{figure}[!t]
	\centering
	\includegraphics[width=3.3in]{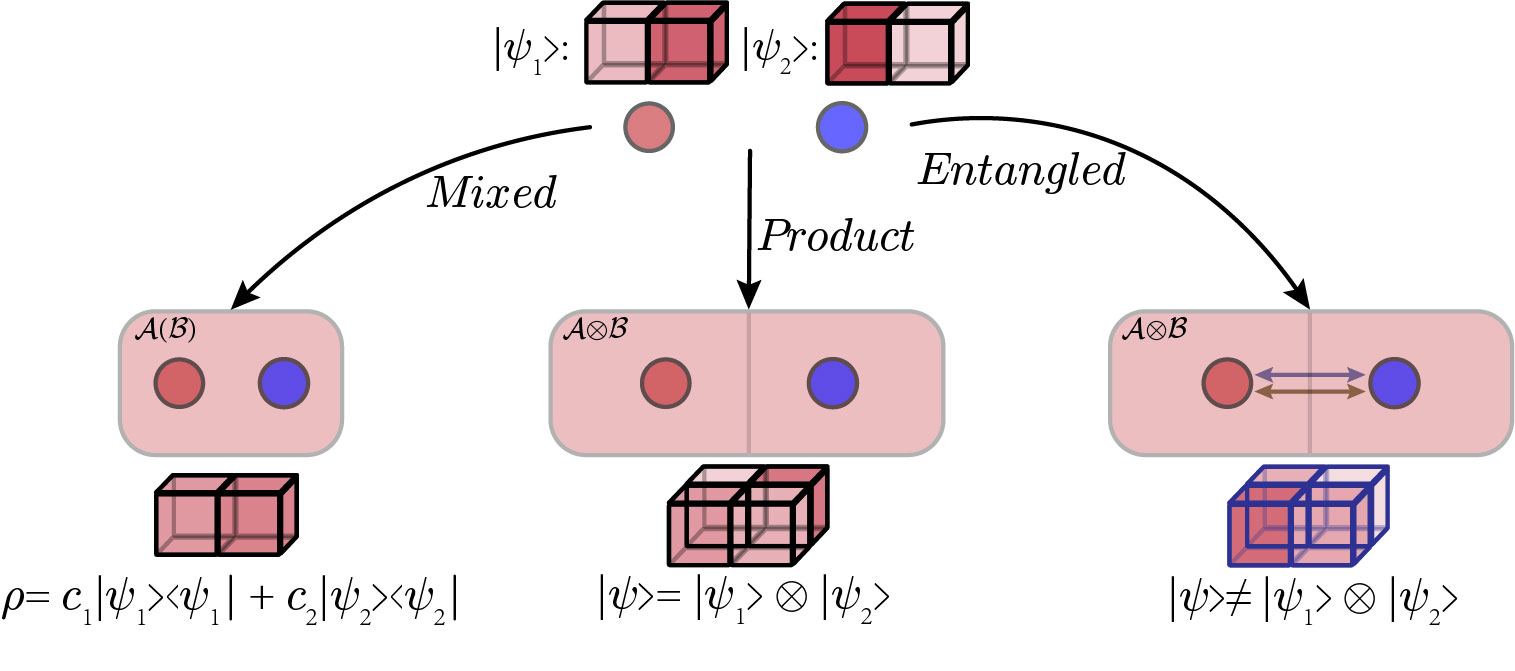}
	\caption{Comparison between mixed state, product state and entangled state with a bipartite example. $|\psi\rangle$ is the system state. $|\psi_1\rangle$ and $|\psi_2\rangle$ are the states of subsystems $\mathcal{A}$ and $\mathcal{B}$, respectively. Parallelized correlations (denoted as arrows) exist between the two subsystems in the entangled state, while for mixed and product states the superposition only exists within the subsystem itself. Separable state is defined as the classical probabilistic mixture of product states. The product state and entangled state are defined on the tensor product $(\otimes)$ of two Hilbert spaces.
	\label{Comparison_of_states}}
\end{figure}

Despite the fact that previous QLMs have achieved good performance and transparency, the state-of-the-art designs still have limitations. For example, the mixed-state representation of the word sequence in \cite{zhang2018end,li2019cnm} is just a classical ensemble of the word states. As shown in the left of Fig.~\ref{Comparison_of_states}, a classical probabilistic mixture of quantum states is not able to fully capture the complex interaction among subsystems. Although Quantum Many-body Wave Function (QMWF) \cite{zhang2018quantum} method has been applied to model the entire word sequence as the combination of subsystems, it is still based on a strong premise that the states of the word sequences are separable, as shown in the middle of Fig.~\ref{Comparison_of_states}. A general quantum state has the ability to describe distributions that cannot be split into independent states of subsystems like in the right of Fig.~\ref{Comparison_of_states}. In quantum physics, the state for a group of particles can be generated as an inseparable whole, leading to a non-classical phenomenon called quantum entanglement \cite{horodecki2009quantum}. Quantum entanglement can be understood as correlations (between subsystems) in superposition, and this type of parallelized correlations can be observed in human language system as well. A word can have different meanings when combined with other words. For example, the verb \textit{turn} has four meanings \textit{\{move, change, start doing, shape on a lathe\}}. If we combine it with \textit{on} to get the phrase \textit{turn on}, the meaning of \textit{turn} will be in the superposition of \textit{change} and \textit{start doing}. However, this kind of correlation has not been explicitly generated in the present NN-based QLMs. Besides, a statistical method has been proposed in \cite{xie2015} to characterize the entanglement within the text in a post-measurement configuration, which still lacks the transparency in the forward modelling process.

In this paper, we propose a novel quantum-inspired Entanglement Embedding (EE) module which can be conveniently incorporated into the present NN-based QLMs to learn the inseparable association among the word states. To be more specific, each word is firstly embedded as a quantum pure state and described by a unit complex-valued vector corresponding to the superposition of sememes. The Word Embedding neural network module is adopted from \cite{li2019cnm}. Word sequences (phrases, $N$-grams, etc.) are initially given as the tensor product of the individual word states, and then transformed to a general entangled state as the output of the EE module. The EE module is realized by a complex-valued neural network, which is essentially approximating the unitary operation that converts the initial product state to an entangled state. After the entanglement embedding, high-level features of the word sequences are extracted by inner products between the entangled state vector and virtual quantum measurement vectors \cite{nielsen2010quantum}. All the parameters of the complex-valued neural network are trainable with respect to a cost function defined on the extracted features. Entanglement measures for quantifying and visualizing the entanglement among the word states can be directly applied on the output of the learned model. We conduct experiments on two benchmark Question Answering (QA) datasets to show the superior performance and post-hoc interpretability of the proposed QLM with EE (QLM-EE). In addition, the word embedding dimension can be greatly reduced when compared with previous QLMs, due to the composition of word embedding and EE modules in the hierarchical structure of the neural network. Note that the current QLM-EE is proposed to implement on classical computing devices with a quantum-inspired neural network structure.

The main contributions of this paper are summarized as follows.
\begin{itemize}
  \item A novel EE neural network module is proposed. The output of the EE module represents the correlations among the word states with a quantum probabilistic model which explores the entire Hilbert space of quantum pure states. The embedded states can reveal the possible entanglement between the words, which is an indication of parallelized correlations. The entanglement can be quantified to promote the transparency and post-hoc interpretability of QLMs to an unprecedented level.
  \item A QLM-EE framework is presented by cascading the word embedding and EE modules. The word embedding module captures the superposed meanings of individual words, while the EE modules encode the correlations between the words at a higher level. The resulting cascaded deep neural network is more expressive and efficient than the shallow networks used by previous QLMs.
  \item The superior performance of QLM-EE is demonstrated over the state-of-the-art classical neural network models and other QLMs on QA datasets. In addition, the word embedding dimension in QLM-EE is greatly reduced and the semantic similarity of the embedded states of word sequences can be studied using the tools from quantum information theory. The entanglement between the words can be quantified and visualized using analytical methods under the QLM-EE framework.
\end{itemize}

This paper is organized as follows. Section II provides a brief introduction to preliminaries and related work. Entanglement embedding is presented in Section III. Section IV proposes the QLM-EE model. Experimental results are presented in Section V and the results show that QLM-EE achieves superior performance over five classical models and five quantum-inspired models on two datasets. Post-hoc interpretability is discussed in Section VI and concluding remarks are given in Section VII.


\section{Preliminaries and related work}

\subsection{Quantum State}
Mathematically, an $n$-level quantum system can be described by an $n$-dimensional Hilbert space $\mathcal{H}=\mathbb{C}^n$. A pure state of the quantum system is described by Dirac notation where a \textit{ket} represents a state vector, written as $\arrowvert\psi\rangle$ (equivalent to a complex-valued column vector). The conjugate transpose, denoted by $\dagger$, of a state vector is called \textit{bra}, denoted as $\langle\psi\arrowvert$, i.e., $\langle\psi\arrowvert=(\arrowvert\psi\rangle)^\dagger$. Denote a chosen orthonormal basis of $\mathcal{H}$ as $\{\arrowvert 0 \rangle, \arrowvert 1 \rangle, \dots, \arrowvert n-1 \rangle\}$. Any quantum pure state can be described by a unit vector in $\mathcal{H}$, which may be expanded on the basis states as
\begin{equation}\label{puredef}
\arrowvert\psi\rangle = \sum_{i = 0}^{n-1} \alpha_{i} \arrowvert i\rangle,
\end{equation}
with complex-valued probability amplitudes $\{\alpha_{i}\}$ satisfying
\begin{equation}
\sum_{i = 0}^{n-1} \arrowvert \alpha_{i} \arrowvert^2 = 1.
\end{equation}
Note that the set $\{|\alpha_{i}|^2\}$ defines a classical discrete probability distribution. A quantum system can be in the superposition of distinct states at the same time, with the probability of being $|i\rangle$ given by $|\alpha_{i}|^2$. For example, we consider a quantum bit (qubit) that is the basic information unit in quantum computation and quantum information, which can be physically realized using e.g., a photon, an electron spin or a two-level atom \cite{nielsen2010quantum}. The state of a qubit can be described as
\begin{equation}
\arrowvert\psi\rangle = \alpha_{0} \arrowvert 0\rangle+\alpha_{1} \arrowvert 1\rangle,
\end{equation}
where $\alpha_{0},\ \alpha_{1}\in \mathbb{C}$ and
\begin{equation}\label{qubitpuredef}
\arrowvert 0\rangle=\begin{bmatrix}
                      1 \\
                      0 \\
                    \end{bmatrix},
\arrowvert 1\rangle=\begin{bmatrix}
                      0 \\
                      1 \\
                    \end{bmatrix}.
\end{equation}

The state of a composite system $\arrowvert\psi_{\mathcal{AB}}\rangle$ consisting of two subsystems $\mathcal A$ and $\mathcal B$ can be described by the tensor product ($\otimes$) of the states of these two subsystems $\arrowvert\psi_{\mathcal{A}}\rangle$ and $\arrowvert\psi_{\mathcal{B}}\rangle$ as
\begin{equation}
\arrowvert\psi_{\mathcal{AB}}\rangle=\arrowvert\psi_{\mathcal{A}}\rangle \otimes \arrowvert\psi_{\mathcal{B}}\rangle.
\end{equation}
For example, if two qubits are in $\arrowvert\psi_1\rangle = \alpha_{0} \arrowvert 0\rangle+\alpha_{1} \arrowvert 1\rangle$ and $\arrowvert\psi_2\rangle = \alpha_{3} \arrowvert 0\rangle+ \alpha_{4} \arrowvert 1\rangle$, respectively, the state of the two-qubit system can be described by
\begin{equation}
\arrowvert\psi_{12}\rangle = \alpha_{0}\alpha_{3} \arrowvert 00\rangle+\alpha_{0}\alpha_{4} \arrowvert 01\rangle+\alpha_{1}\alpha_{3} \arrowvert 10\rangle+\alpha_{1}\alpha_{4} \arrowvert 11\rangle,
\end{equation}
where we have denoted
\begin{equation}\nonumber
\arrowvert 0\rangle\otimes \arrowvert 0\rangle=\arrowvert 00\rangle=\begin{bmatrix}
                      1 \\
                      0 \\
                      0 \\
                      0 \\
                    \end{bmatrix},
\arrowvert 0\rangle\otimes \arrowvert 1\rangle=\arrowvert 01\rangle=\begin{bmatrix}
                      0 \\
                      1 \\
                      0 \\
                      0 \\
                    \end{bmatrix},
\end{equation}
\begin{equation}\nonumber
\arrowvert 1\rangle\otimes \arrowvert 0\rangle=\arrowvert 10\rangle=\begin{bmatrix}
                      0 \\
                      0 \\
                      1 \\
                      0 \\
                    \end{bmatrix},
\arrowvert 1\rangle\otimes \arrowvert 1\rangle=\arrowvert 11\rangle=\begin{bmatrix}
                      0 \\
                      0 \\
                      0 \\
                      1 \\
                    \end{bmatrix}.
\end{equation}

For an open quantum system or a quantum ensemble, its state needs to be described by a density matrix $\rho$ satisfying $\text{tr}(\rho)=1$, $\rho^\dagger=\rho$ and $\rho\geq 0$. In this paper, we mainly focus on quantum pure states, and thus the inputs and outputs of the neural network modules are complex-valued vectors that stand for the pure states. If a quantum system is in the state in (\ref{puredef}), then the system is physically in the superposition state of $\{\arrowvert i\rangle\}$.
Similar superposition, although not physically, may also exist in the human language systems, which is expressed as the superposition of multiple meanings of a semantic unit.

\subsection{Quantum Entanglement}
Quantum entanglement is one of the most fundamental concepts in quantum theory. Entanglement describes the non-classical correlation between quantum systems. To be more specific, a many-body quantum system is in an entangled state if the state of one subsystem is determined by the measurement result of the other subsystem. Mathematically speaking, the joint state of an entangled quantum system cannot be decomposed into the states of subsystems by tensor product. For example, we consider two quantum systems $\mathcal A$ and $\mathcal B$ defined in Hilbert spaces $\mathcal{H_{A}}$ and $\mathcal{H_{B}}$, respectively. Assume the basis state vectors of the two subsystems are $\{\arrowvert i \rangle\}$ and $\{\arrowvert j \rangle\}$. The joint state is then defined on the tensor product space $\mathcal{H_{A}}\otimes\mathcal{H_{B}}$, whose basis state vectors are given by the set $\{\arrowvert i \rangle\otimes\arrowvert j \rangle\}$. A general pure state $\arrowvert \psi \rangle$ of the composite quantum system can be written as follows
\begin{equation}
\arrowvert \psi \rangle = \sum_{i,j}^{n} \alpha_{ij}\arrowvert i \rangle \otimes \arrowvert j \rangle,
\label{Joint_state}
\end{equation}
where $\{\alpha_{ij}\}$ are complex-valued probability amplitudes. The pure state is \textbf{separable} if it can be decomposed as
\begin{equation}
\arrowvert \psi \rangle = \arrowvert \psi_{1} \rangle \otimes \arrowvert \psi_{2} \rangle, \label{Separable_state}
\end{equation}
where $\arrowvert \psi_{1} \rangle = \sum_{i}\alpha_{1i}\arrowvert i \rangle$ and $ \arrowvert \psi_{2}\rangle = \sum_{j}\alpha_{2j} \arrowvert j \rangle$ are pure states of the subsystems. Otherwise, the pure state is \textbf{entangled}. According to (\ref{Joint_state}) and (\ref{Separable_state}), separable pure states only constitute a small potion of the quantum states that can be defined on $\mathcal{H_{A}}\otimes\mathcal{H_{B}}$, which means that a significant amount of correlations between the subsystems cannot be characterized by the separable states. For example, one of the entangled \emph{Bell States} or \emph{EPR pairs} \cite{nielsen2010quantum} is defined by
\begin{equation}
\arrowvert \psi \rangle = \frac{1}{\sqrt{2}} ( \arrowvert 00 \rangle + \arrowvert 11 \rangle),
\end{equation}
which can not be written as a tensor product of two pure states of the subsystems. The composite system is in the superposition of two basis states $\arrowvert 00 \rangle$ and $\arrowvert 11 \rangle$. If we measure the state of the first system and the measurement result is $|1\rangle$, then the state of the second system is $|1\rangle$. However, since the first system is a superposition of two states $|0\rangle$ and $|1\rangle$, the measurement result can be $|0\rangle$ with equal probability. In that case, the state of the second system is $|0\rangle$. This kind of non-classical correlation cannot be modelled by classical probability. Quantum entanglement can be used to model the superposition of correlations between the subsystems, or in our case, the superposition of multiple meanings between the word states.

\begin{figure*}[!t]
	\centering
	\includegraphics[width=5in]{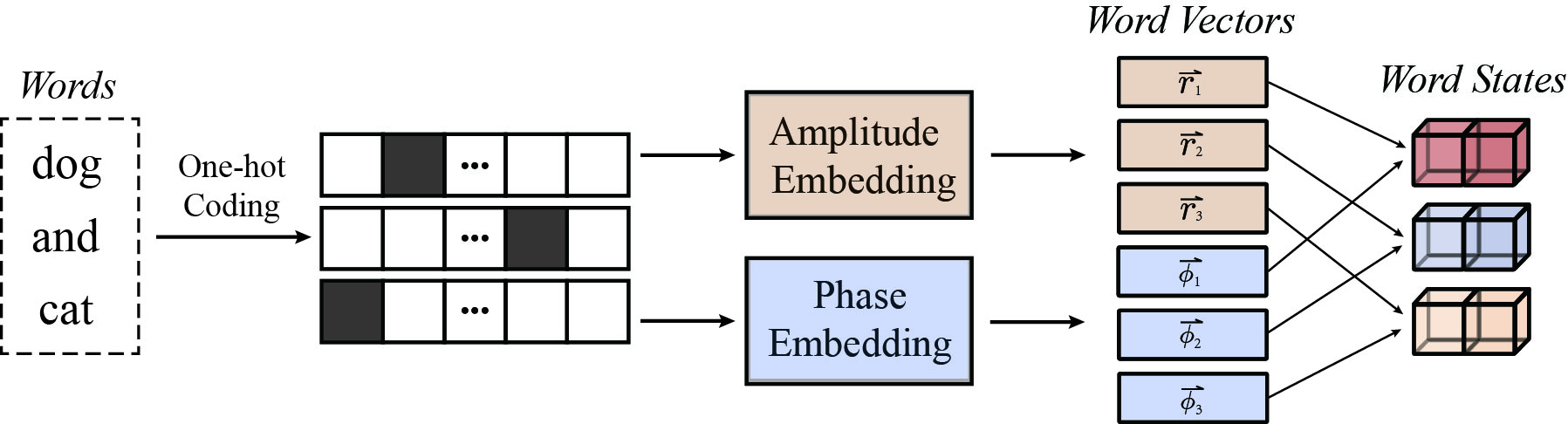}
	\caption{Pipeline of the word embedding module for a 3-word sequence.}
	\label{Word_embedding}
\end{figure*}

\subsection{Quantum Measurement}
Quantum measurement is used to extract information from a quantum system. A widely-used measurement is the projective measurement (von Neumann measurement). For example, when measuring a pure state $\arrowvert\psi\rangle = \sum_{i = 0}^{n-1} \alpha_{i} \arrowvert i\rangle$ by projecting onto the measurement basis states $\{\arrowvert i\rangle\}$, the quantum state will collapse to one of the basis states with the probability of
\begin{equation}\label{qmdef}
p_{i}(\arrowvert \psi\rangle) = \arrowvert \alpha_{i} \arrowvert^2 =  |\langle i\arrowvert \psi\rangle|^2,
\end{equation}
and the inner product $\langle i\arrowvert \psi\rangle$ of $\arrowvert i\rangle$ and $\arrowvert \psi\rangle$ is calculated as
\begin{equation}
\langle i\arrowvert \psi\rangle= (\arrowvert i\rangle)^\dagger \arrowvert \psi\rangle.
\end{equation}
In a more general setting, projective measurements can be performed using any state vector $\arrowvert x \rangle$ (i.e., not just the computational basis), with the probability of obtaining $|x\rangle$ given by
\begin{equation}
p_{x}(\arrowvert \psi\rangle) =|\langle x\arrowvert \psi\rangle|^2.
\label{Measure}
\end{equation}

\subsection{Quantum Fidelity}
In quantum information theory, fidelity is a real-valued measure of the similarity between two quantum pure states, which is defined as
\begin{equation}
\mathcal F(|\psi_a\rangle,|\psi_b\rangle)=|\langle\psi_a|\psi_b\rangle|^2.
\end{equation}
According to (\ref{qmdef}), fidelity is just the probability of collapsing $|\psi_a\rangle$ to $|\psi_b\rangle$ if $|\psi_a\rangle$ is measured by $|\psi_b\rangle$, or the probability of collapsing $|\psi_b\rangle$ to $|\psi_a\rangle$ if $|\psi_b\rangle$ is measured by $|\psi_a\rangle$. In other words, fidelity is the probability that one quantum state will pass the test to be identified as the other.

\subsection{Complex-valued Word Embedding} \label{we}
Complex-valued word embedding module aims to model words as quantum pure states in the semantic Hilbert space. In this paper, the complex-valued word embedding module is adopted from \cite{li2019cnm}. As shown in Fig.~\ref{Word_embedding}, each word is firstly encoded to a one-hot vector with a fixed length. Then, the amplitude embedding and phase embedding modules map the one-hot vector into a pair of real-valued amplitude and phase vectors $\{[r_{1} \cdots r_{n}]^{T}, [\phi_{1} \cdots \phi_{n}]^{T}\}$.  After that, the amplitude vector is normalized to a unit vector and the polar form representation of the word state is given by
	\begin{equation}\label{word_polar}
	|s\rangle=\sum_{j=1}^{n} r_{j} e^{\rm i \phi_{j}}\left|j\right\rangle,
	\end{equation}
where $\rm i$ is the imaginary number with $\rm i^2 = -1$. Note that $\left\{\left|j\right\rangle\right\}_{j=1}^{n}$ are the basis sememes of the semantic Hilbert space, which represent the minimum semantic units of the word meaning. Finally, the pair of vectors is transformed into a complex-valued vector as $[\alpha_{1} \cdots \alpha_{n}]^{T}$, and the word state can be written as
	\begin{equation}\label{word}
	|s\rangle=\sum_{j=1}^{n} \alpha_{j} |j\rangle
	\end{equation}
with $\sum_{j=1}^{n}|\alpha_j|^2=1$. The complex-valued word embedding in \cite{wang2019encoding} defines the real-valued amplitude as the semantic meaning of the word, and the complex phases as the positional information of word in the sequence. In contrast, the complex-valued word embedding in this paper aims to model words using quantum state representation, and no specific meaning is given to the phase or amplitude. Instead, the semantic meanings and their quantum-like superposition are jointly determined by the amplitude and phase.

\subsection{Related Work}
In \cite{van2004geometry}, van Rijsbergen argued that quantum theory can provide a unified framework for the geometrical, logical and probabilistic models for information retrieval. Coecke et al. \cite{coecke2010mathematical} introduced DisCo formalism based on tensor product composition and Zeng et al. \cite{zeng2016quantum} presented a quantum algorithm to categorize sentences in DisCo model. Kartsaklis \cite{Kartsaklis2014entangle} used the traditional machine learning method to quantify entanglement between verbs in DisCo model. Sordoni et al. \cite{sordoni2013modeling} proposed a quantum language modelling approach for information retrieval and the density matrix formulation was used as a general representation for texts. The single and compound terms were mapped into the same quantum space, and term dependencies are modelled as projectors. In \cite{sordoni2014learning}, Quantum entropy minimization method has been proposed in learning concept embeddings for query expansion, where concepts from the vocabulary were embedded in rank-one matrices, and documents and queries were described by the mixtures of rank-one matrices. In \cite{basile2017towards}, a quantum language model was presented where a ``proof-of-concept" study was implemented to demonstrate its potential.

Two NN-based Quantum-like Language Models (NNQLMs) have been proposed, namely NNQLM-\uppercase\expandafter{\romannumeral1} and NNQLM-\uppercase\expandafter{\romannumeral2} \cite{zhang2018end}. Words were modelled as quantum pure states in the semantic Hilbert space and a word sequence was also modelled in the same space by mixing the word states in a classical way as
\begin{equation}\label{ms}
	\rho=\sum_{k} c_k\left| s_{k}\right\rangle\left\langle s_{k}\right|,
\end{equation}
where $\arrowvert s_{k}\rangle$ was the word state representing the $k$-th word in the sentence and $c_{k}$ is the weight of $k$-th word state satisfying $\sum_{k}c_{k}=1$. By (\ref{ms}), the semantic meaning of the word sequence is mainly determined by the word states with larger weights. In NNQLM-\uppercase\expandafter{\romannumeral1}, the representation of a single sentence was obtained by the first three layers as a density matrix corresponding to a mixed state, and then the joint representation of a question/answer pair was generated in the fourth layer by matrix multiplication. The last softmax layer was invoked to match the question/answer pair. NNQLM-\uppercase\expandafter{\romannumeral2} adopted the same first 4-layer network structure to obtain the joint representation of a question/answer pair as NNQLM-\uppercase\expandafter{\romannumeral1}, and employed a 2-dimensional convolutional layer instead to extract the features of the joint representation for comparing the question/answer pairs. In \cite{zhang2018quantum}, a Quantum Many-body Wave Function (QMWF) method was presented in which the representation of a single sentence was given by the tensor product of word vectors. Operation that mimics the quantum measurement was applied on the product state to extract the correlation patterns between the word vectors. To be more specific, a three-layer Convolutional Neural Network (CNN) was used, in which the first layer generated the product state representation of a word sequence and the projective measurement on the product state was simulated by a 1D convolutional layer with product pooling. A complex-valued network called CNM was presented in \cite{li2019cnm}. Similar to NNQLMs, CNM embedded the word sequence as a mixed state but with a complex-valued neural network. Then a number of trainable measurement operations were applied on the complex-valued density matrix representations to obtain the feature vectors of question/answer for comparison. CNM achieved comparable performance over the state-of-the-art models based on CNN and recurrent NN. More importantly, CNM has shown its advantage in interpretability, since the model has simulated the generation of a quantum probabilistic description for the individual words with complex-valued word states, and the projection of the superposed sememes onto a fixed meaning by quantum-like measurement within a particular context. The work \cite{Uprety2020survey} presented a survey of the quantum-inspired information retrieval and drew a road-map of future directions. Ref. \cite{Gkoumas2021quantum} proposed a decision-level fusion strategy for predicting sentiment judgments inspired by quantum theory.

In the existing works, some potential of quantum-inspired NN modules have been explored for language modelling. However, most of the existing works assume that the word states are placed in the same Hilbert space, while in QLM-EE the word states are placed in the tensor product of Hilbert spaces. This enables an explicit entanglement analysis between the word states, which is not possible if the word states are mixed in the same space. Although \cite{zhang2018quantum} has modelled the words in independent Hilbert spaces, it was assumed that the interaction among the words are described by product states, which limits the expressive power of quantum state space. This paper continues these efforts and proposes a novel entanglement embedding module for question answering task.

\section{Entanglement Embedding}
Quantum probabilistic superposition of states models the polysemy of words and sequences. Quantum states span the entire complex Hilbert space, while the amplitudes and complex phases of the states give the probability distribution of the multiple meanings of words. An $N$-gram is the composition of $N$ words whose joint state is defined on the tensor product of $N$ Hilbert spaces. However, the joint state itself is not necessarily the tensor product of $N$ states of the Hilbert spaces. The set of entangled state is significantly larger than that of the tensor-product states, and compound semantic meaning lying within the $N$-gram is mainly captured by the entanglement representation. Then the parameters of the entangled states is trained by the EE module, which characterizes the features of entanglement between the word states within the sequence.

The EE module is trained to capture high-level features between the words in the $N$-gram, while the word embedding module can focus on encoding the basic semantic meanings of individual words with reduced dimension. The clear separation of duties in the hierarchical structure of the neural network increases the transparency of the model to an unprecedented level, which allows an accurate interpretation of the intermediate states using the tools borrowed from quantum information theory.

In line with the previous works, the complex-valued word embedding module is used to transform the one-hot representation of the word into a quantum pure state in the Hilbert space of word states $\mathcal{H}_w$, expressed as a state vector $|\psi\rangle=[\alpha_{0}\ \cdots\ \alpha_{i}\ \cdots]^T$, with $\sum_i|\alpha_i|^2=1$. The general quantum pure state for describing a word sequence is defined on the tensor-product Hilbert space $\mathcal{H}_s := \otimes_i(\mathcal{H}_w)_i$, and can be formulated as
\begin{equation}
\arrowvert \psi_s \rangle =  \sum_{i_1,\dots,i_N} \beta_{i_1\dots i_N} \arrowvert i_1 \rangle \otimes \cdots \otimes \arrowvert i_N \rangle,
\label{General_state_joint}
\end{equation}
where $\sum_{i_1,\dots,i_N} |\beta_{i_1\dots i_N} |^2= 1$. $\arrowvert \psi_s \rangle$ is used to characterize the probability distribution of the sememes. If the word embedding dimension is $D$, then a general pure state vector defined on the tensor product of $N$ Hilbert spaces contains $D^N$ elements.

\begin{figure}[!t]
	\centering
	\includegraphics[width=3.3in]{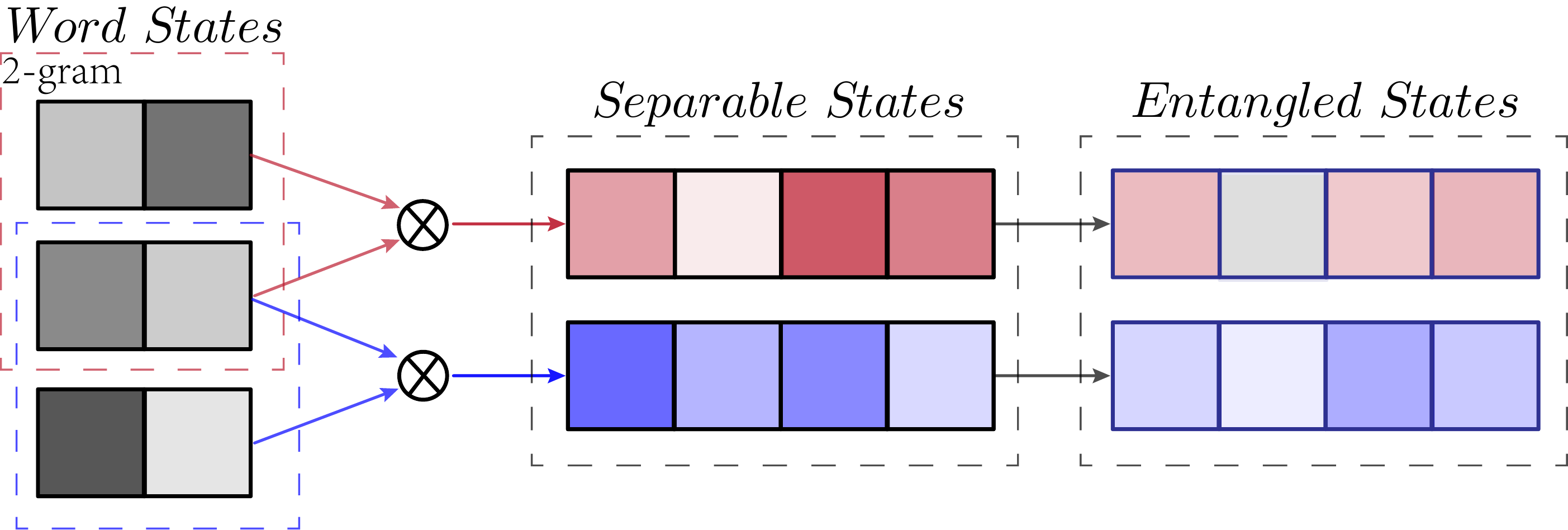}
	\caption{Pipeline of the EE module for a 3-word sequence. $\otimes$ denotes the tensor product of input vectors.}
	\label{Ent_embedding}
\end{figure}

\begin{figure*}[!t]
	\centering
	\includegraphics[width=7in]{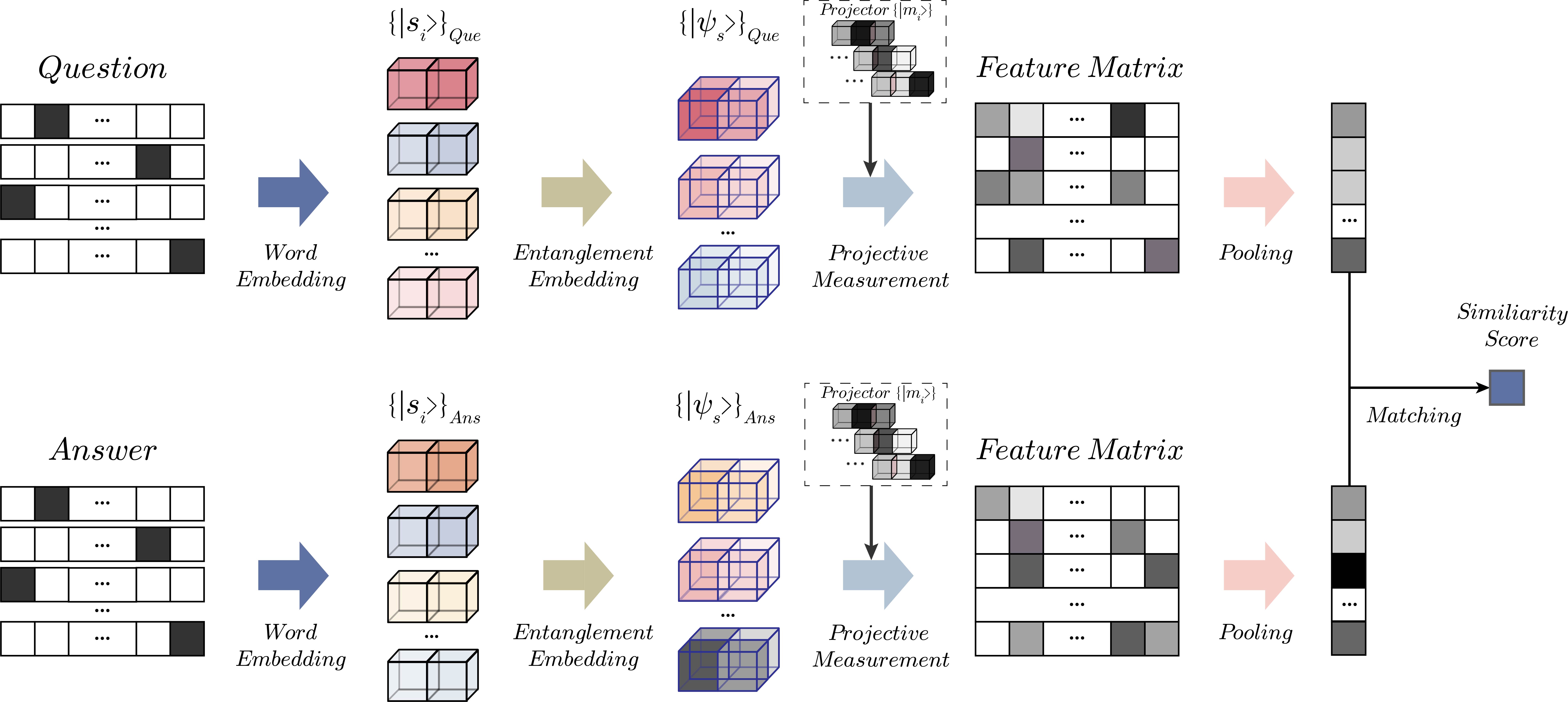}
	\caption{The structure of QLM-EE.}
	\label{Structure}
\end{figure*}

The EE module starts with composing the contiguous sequences of words as $N$-grams \cite{cavnar1994n}; see Fig.~\ref{Ent_embedding} for an example. For each $N$-gram, a separable pure state is generated as the tensor product of its word states taking the following form
\begin{equation} \label{tensor_product}
\arrowvert \psi_{ss} \rangle =  \arrowvert \psi_1 \rangle \otimes \cdots \otimes \arrowvert \psi_N \rangle.
\end{equation}
The word states are defined on Hilbert spaces which are isomorphic to each other, with the same semantic basis states. For this reason, $N$-grams are defined on the same tensor-product space before entering the EE module. The complex-valued EE module is then connected to transform the initial separable state to an unnormalized vector $\arrowvert \tilde{\psi}_s \rangle=[\tilde{\beta}_{i_1\dots i_N}]$ which will then be normalized to determine a general pure state $\arrowvert \psi_s \rangle=[\beta_{i_1\dots i_N}]$ in the form of (\ref{General_state_joint}). The transformation induced by the NN can be formally written as
\begin{equation}
\arrowvert \tilde{\psi}_s \rangle=\mathcal W\arrowvert \psi_{ss} \rangle,\label{weightmatrix}
\end{equation}
where $\mathcal W$ is the weight matrix. Then the output vector must be normalized by
\begin{equation}
\beta_{i_1\dots i_N}=\frac{\tilde{\beta}_{i_1\dots i_N}}{\sum_{i_1,\dots,i_N} |\beta_{i_1\dots i_N} |^2}
\label{Normalize}
\end{equation}
to be consistent with quantum theory. Note that the operations (\ref{weightmatrix})-(\ref{Normalize}) induce an endomorphism of this Hilbert space and the EE module transfers one pure state to another. The pure states are unit vectors in the tensor-product Hilbert space. Any unit vectors in the same Hilbert space can be connected by a unitary matrix which induces a rotation. It is known that for any two quantum pure states $|a\rangle$ and $|b\rangle$ of an $N$-qubit register, there exists a gate sequence to physically realize the unitary matrix $U$ such that $|b\rangle=U|a\rangle$ \cite{onen2005transformation}. Since the model proposed in this paper is only quantum-inspired, we are using a linear matrix multiplication and a normalization layer to approximate the effect of the transformation matrix by the classical way of training. In principle, if one layer of linear matrix multiplication is not sufficient for learning the transformation, it is always possible to stack several layers of linear operations to accurately approximate any transformation, which is consistent with the traditional neural network theory.

We take Fig.~\ref{Ent_embedding} as an example to illustrate the working mechanism of the EE module. We denote the state vectors for the first two words as $[\alpha_{1}\ \alpha_{2}]^T$ and $[\alpha_{3}\ \alpha_{4}]^T$. The two word states form the separable state as the input to the NN layer by the following tensor product
\begin{equation}\left[\begin{array}{l}
\alpha_{1} \\
\alpha_{2}
\end{array}\right] \otimes\left[\begin{array}{l}
\alpha_{3} \\
\alpha_{4}
\end{array}\right]=\left[\begin{array}{l}
\alpha_{1} \alpha_{3} \\
\alpha_{1} \alpha_{4} \\
\alpha_{2} \alpha_{3} \\
\alpha_{2} \alpha_{4}
\end{array}\right].
\label{Sep_coe_vec}
\end{equation}
The output of the NN layer is an unnormalized vector $[\tilde{\beta}_{00}\ \tilde{\beta}_{01}\ \tilde{\beta}_{10}\ \tilde{\beta}_{11}]^T$. After normalization, we obtain
\begin{equation}\left[\begin{array}{c}
\beta_{00} \\
\beta_{01} \\
\beta_{10} \\
\beta_{11}
\end{array}\right],
\label{Ent_coe_vec}
\end{equation}
which is in the most general form of the state vector on the joint Hilbert space. If $[\beta_{00}\ \beta_{01}\ \beta_{10}\ \beta_{11}]^T$ cannot be written in a decomposable form just like the RHS of (\ref{Sep_coe_vec}), then the output vector is a representation for an entangled state which captures the non-classical correlations between the word states.

\begin{figure*}[t]
	\centering
	\includegraphics[width=6.5in]{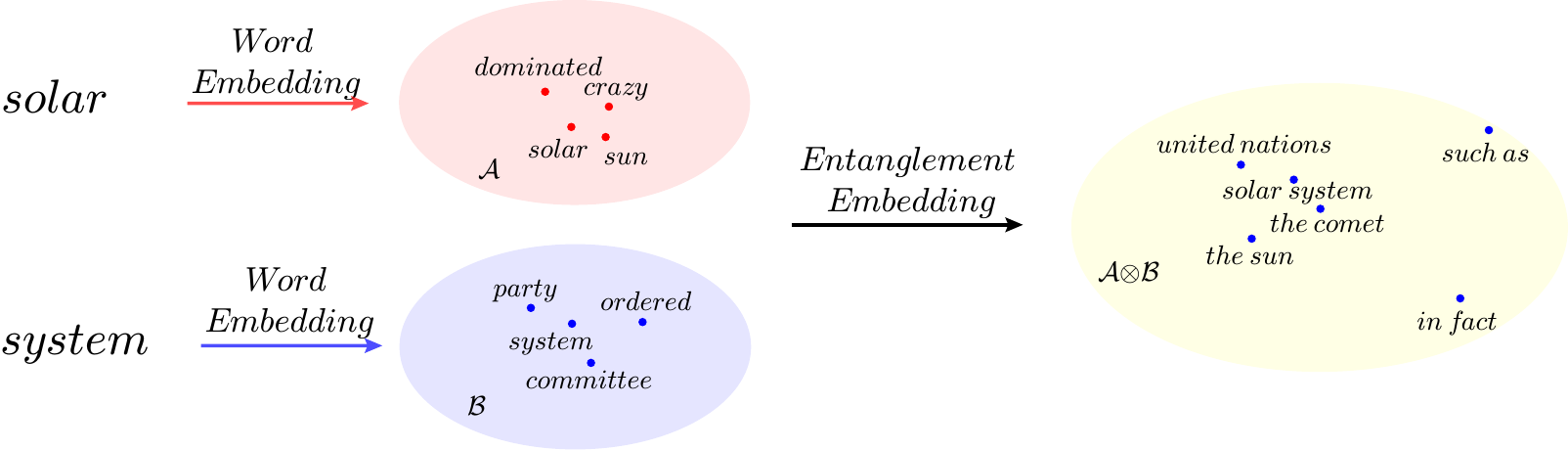}
	\caption{The forward process for the phrase \textit{solar system}. The distance between state vectors is defined by the quantum fidelity. The fidelities of (\textit{solar system}, \textit{united nations}), (\textit{solar system}, \textit{the comet}), (\textit{solar system}, \textit{the sun}) are 0.523, 0.365, 0.335, respectively, while the fidelities of randomly selected combinations (\textit{solar system}, \textit{in fact}), (\textit{solar system}, \textit{such as}) are 0.026, 0.015, respectively.}
	\label{Example4EE}
\end{figure*}


\section{Model}
 The structure of QLM-EE for QA is shown in Fig.~\ref{Structure}. It is a complex-valued, end-to-end neural network optimized by back-propagation on classical computing devices. The QLM-EE can be divided into three major steps.
\begin{itemize}
	\item \textit{Word embedding and entanglement embedding.} The word embedding module generates the word states, and the entanglement embedding module generates the complete quantum probabilistic description of the sequence of word states. The complete probabilistic description is given by a generic quantum pure state vector, which may be entangled to encode the information on the non-classical correlations between the word states. Entanglement embedding and word embedding modules encode the states as feature vectors at different levels, which improves the transparency of the quantum probabilistic modelling process. In particular, the distance between the feature vectors in the embedding space can be calculated based on the well-established measures from the quantum information theory for comparing quantum states, which could be used to reveal the relations between the words and phrases on a deeper semantic level. For example, in the classical Word2Vec, the cosine similarity is defined on real-valued embedded vectors as
\begin{equation}\label{cosreal}
\mathcal C(\vec{a},\vec{b})=\frac{\vec{a}\cdot\vec{b}}{||\vec{a}|| \cdot ||\vec{b}||}.
\end{equation}
In our case, the cosine similarity is defined on the complex-valued unit vectors as
\begin{equation}\label{coscomplex}
\mathcal C(\vec{a},\vec{b})=\frac{\vec{a}^\dagger\vec{b}}{||\vec{a}|| \cdot ||\vec{b}||}=\vec{a}^\dagger\vec{b}.
\end{equation}
In general, $\mathcal C(\vec{a},\vec{b})$ is a complex number, which means that the state representations could differ by a complex phase factor. Since complex phases are hard to visualize, we employ the quantum fidelity measure to compare the words and word sequences.

In Fig.~\ref{Example4EE}, we visualize the process how the state of the phrase \textit{solar system} evolves in our model using the quantum fidelity measure. After the word embedding layer, the state vector that represents the word \textit{solar} is close to $\{$\textit{sun}, \textit{crazy}, \textit{dominated}$\}$, which reflects how the model comprehends the meaning of the input words. The word state vector of \textit{system} is embedded close to $\{$\textit{party}, \textit{committee}, \textit{ordered}$\}$. After the EE module, the phrase \textit{solar system} can be linked to other high-level phrases, e.g., \textit{united nations}, while its state vector is still very close to the phrase \textit{the sun} and \textit{the comet} which share a similar meaning.

	\item \textit{Measurement operation.} Projective measurement operation is equivalent to the calculation of fidelity between quantum states. That is, the output of the measurement can be seen as a distance between the measured state and a measuring state, corresponding to the distance between the semantic entangled state and a measuring sememe which is used for comparison. Therefore, performing the same set of measurement operations provides a way to compare the semantic similarity of the question and answering word sequences. After entanglement embedding, a series of parameterized measurements $\{\arrowvert m_i \rangle\}$ are performed on the state $\arrowvert \psi_{s} \rangle$ via the formula
	\begin{equation}
	p_{i}(\arrowvert \psi_{s} \rangle) =  |\langle m_i\arrowvert \psi_{s}\rangle|^2.
	\end{equation}
	Here $p_{i}$ is defined as the measurement output, which indicates the probability of the state $\arrowvert \psi_{s} \rangle$ possessing the semantic meaning represented by the measurement vector $\arrowvert m_i \rangle$. Note that the unit vectors $\{\arrowvert m_i \rangle \}$ are optimized in a data-driven way. By using pure states as measurement basis, the computation cost for measuring a quantum state virtually is reduced from $O(n^3)$ to $O(n)$ compared to CNM \cite{li2019cnm}, in which density matrices were used as the measurement basis. For a sentence described by the concatenation of $L$ word sequences, the feature matrix which stores all the measurement results has $L \times M$ entries if $M$ measurement vectors are used.
	
	\item \textit{Similarity Measure.} A max-pooling layer is applied on each row of the feature matrix for down-sampling. To be more specific, the down-sampling takes the maximum value of each row of the matrix to form a reduced feature vector. Then a vector-based similarity metric can be employed in the matching layer for evaluating the distance between the pair of feature vectors for question and answer. The answer with the highest matching score is chosen as the predicted result among all candidate answers.
\end{itemize}


\section{Experiment} \label{Experiment}

\subsection{Experiment Details}
\begin{table}[!t]
	\renewcommand{\arraystretch}{1.2}
	\caption{statistics of the datasets}
	\label{Statistics_Datasets}
	\centering
	\begin{tabular}{l|rrr}
		\toprule[1pt]
		\textbf{Dataset}&\textbf{Train(Q/A)}&\textbf{Dev(Q/A)}&\textbf{Test(Q/A)}\\ \hline
		TREC-QA&1,229/53,417&65/1,134&68/1,478 \\
		$\rm W_{IKI}$QA&873/8,627&126/1,130&243/2,351  \\
		\toprule[1pt]
	\end{tabular}
\end{table}

\subsubsection{Dataset}
We conduct the experiments on two benchmark datasets for QA, namely TREC-QA \cite{voorhees2000building} and $\rm W_{IKI}$QA \cite{yang2015wikiqa}. TREC-QA is used in the Text REtrieval Conference. $\rm W_{IKI}$QA is an open-domain QA dataset released by Microsoft Research. The statistics of the datasets are given in TABLE~\ref{Statistics_Datasets}.

\subsubsection{Evaluation metrics}
The metrics called Mean Average Precision (MAP) and Mean Reciprocal Rank (MRR) \cite{manning2008introduction} are utilized to evaluate the performance of the models. MAP for a set of queries $Q$ is the mean of the Average Precision scores $\mathrm{AveP(q)}$ for each query, formulated as $$\mathrm{MAP}=\sum_{q=1}^{|Q|} \mathrm{AveP}(\mathrm{q})/|Q|.$$ MRR is the average of the Reciprocal Ranks of results for a sample of queries $Q$, calculated as $$\mathrm{MRR}=\frac{1}{|Q|} \sum_{i=1}^{|Q|} 1/{\mathrm{rank}_{i}},$$ where $\mathrm{rank}_{i}$ refers to the rank position of the first relevant document for the $i$-th query.

\subsubsection{Baselines}

\begin{enumerate}[a)]
	\item Classical models for TREC-QA including
\begin{itemize}
  \item Unigram-CNN \cite{yu2014deep}: Unigram-CNN is a CNN-based model that utilizes 1-gram as the input to obtain the representations of questions and answers for comparison. It is composed of one convolutional layer and one average pooling layer.
  \item Bigram-CNN \cite{yu2014deep}: Bigram-CNN has the same network structure as Unigram-CNN but it extracts the representations from bi-gram inputs.
  \item ConvNets \cite{severyn2015learning}: ConvNets is built upon two distributional sentence models based on CNN. These underlying sentence models work in parallel to map questions and answers to their distributional vectors, which are then used to learn the semantic similarity between them.
  \item QA-LSTM-avg \cite{tan2015lstm}: QA-LSTM-avg generates distributed representations for both the question and answer independently by bidirectional LSTM outputs with average pooling, and then utilizes cosine similarity to measure their distance.
  \item aNMM-1 \cite{yang2016anmm}: aNMM-1 employs a deep neural network with value-shared weighting scheme in the first layer, which is followed by fully-connected layers to learn the sentence representation. A question attention network is used to learn question term importance and produce the final ranking score.
\end{itemize}
\item Classical models for $\rm W_{IKI}$QA including
 \begin{itemize}
   \item Bigram-CNN \cite{yu2014deep};
   \item PV-Cnt \cite{yang2015wikiqa}: PV-Cnt is the Paragraph Vector (PV) model combined with Word Count. The model score of PV is the cosine similarity score between the question vector and the sentence vector.
   \item CNN-Cnt \cite{yang2015wikiqa}: CNN-Cnt employs a Bigram-CNN model with average pooling and combines it with Word Count.
   \item QA-BILSTM \cite{santos2016attentive}: QA-BILSTM uses a bidirectional LSTM and a max pooling layer to obtain the representation of questions and answers, and then computes the cosine similarity between the two representations.
   \item LSTM-attn\cite{miao2016neural}: LSTM-attn firstly obtains the representations for the question and answer from independent LSTM models, and then adds an attention model to learn the pair-specific representation for prediction on the basis of the vanilla LSTM.
 \end{itemize}
	\item Quantum models for TREC-QA and $\rm W_{IKI}$QA including
QLM-MLE \cite{sordoni2013modeling}, NNQLM-\uppercase\expandafter{\romannumeral1}, NNQLM-\uppercase\expandafter{\romannumeral2} \cite{zhang2018end}, QMWF-LM \cite{zhang2018quantum}, and CNM \cite{li2019cnm}. These models have been briefly introduced in Section II.E.
\end{enumerate}

\begin{figure}[!t]
	\centering
	\includegraphics[width=3.3in]{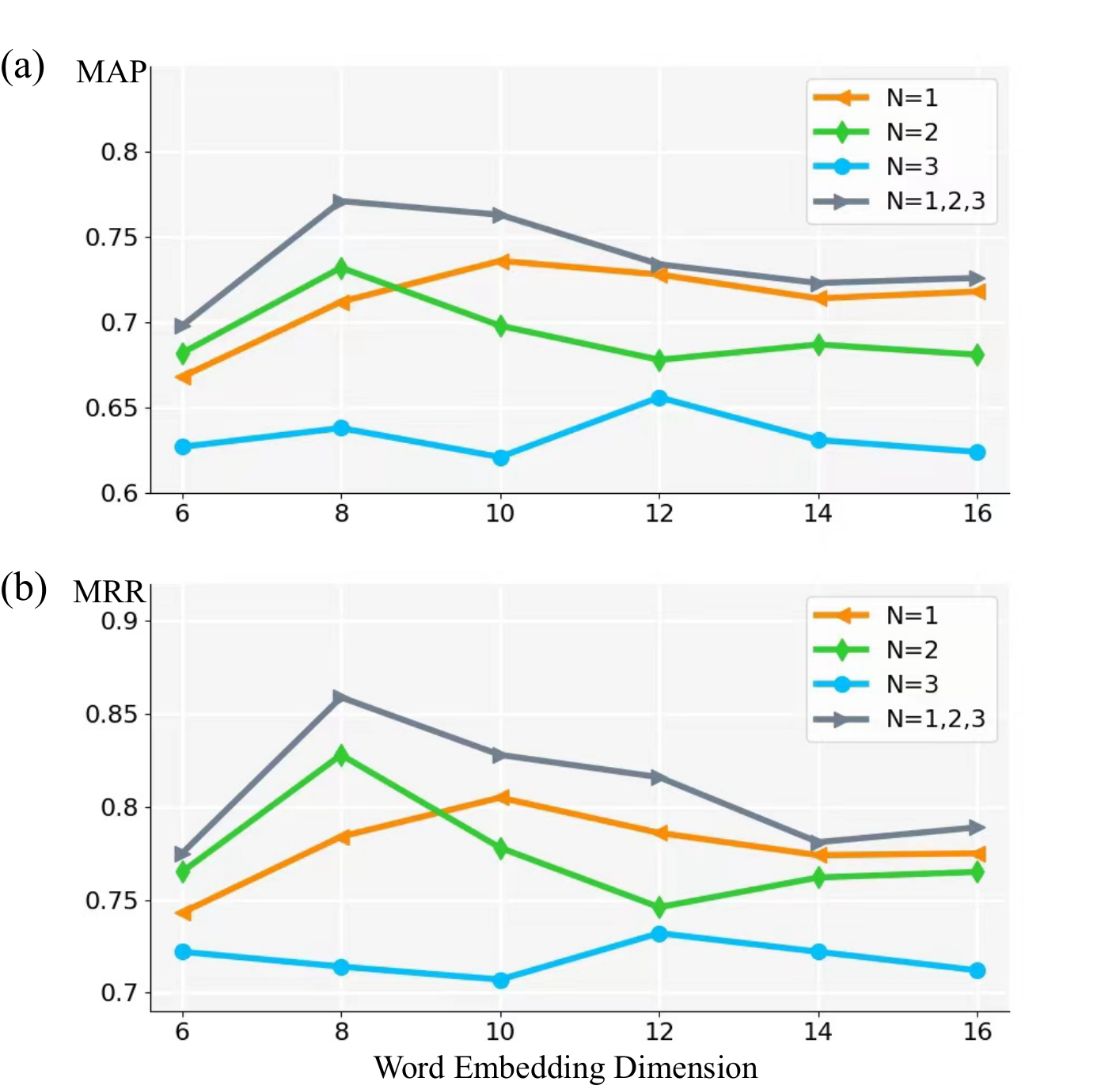}
	\caption{Experimental results of models with different $N$ and different word embedding dimensions on TREC-QA. $N=1,2,3$ refers to the model in which 1-gram, 2-gram and 3-gram entanglement embeddings are applied in parallel on the questions and answers, and the feature vectors are concatenated as a single vector for comparison.}
	\label{fig_sim}
\end{figure}

\begin{table}[!t]
	\renewcommand{\arraystretch}{1.2}
	\caption{experimental results on TREC-QA and $\rm W_{IKI}$QA}
	\label{Result}
	\centering
	\begin{tabular}{l|ll|l|ll}
		\toprule[1pt]
		\multicolumn{3}{c|}{TREC-QA} & \multicolumn{3}{c}{$\rm W_{IKI}$QA}\\
		\hline
		\textbf{Model} & \textbf{MAP}  & \textbf{MRR} & \textbf{Model} & \textbf{MAP}  & \textbf{MRR} \\
		\hline
		Unigram-CNN  & 0.5470 & 0.6329 & Bigram-CNN  & 0.6190 & 0.6281\\
		Bigram-CNN  & 0.5693 & 0.6613 &PV-Cnt &  0.5976  & 0.6058\\
		ConvNets  & 0.6709 & 0.7280 &CNN-Cnt  & 0.6520  & 0.6652\\
		QA-LSTM-avg  & 0.6819  & 0.7652 &QA-BILSTM  & 0.6557 & 0.6695\\
		aNMM-1  & 0.7385 & 0.7995 &LSTM-attn  & 0.6639 & 0.6828\\
		\toprule[1pt]
		QLM-MLE  & 0.6780 & 0.7260 &QLM-MLE  & 0.5120 & 0.5150\\
		NNQLM-\uppercase\expandafter{\romannumeral1}  & 0.6791 & 0.7529 &NNQLM-\uppercase\expandafter{\romannumeral1}  & 0.5462 & 0.5574\\
		NNQLM-\uppercase\expandafter{\romannumeral2}  & 0.7589  & 0.8254  &NNQLM-\uppercase\expandafter{\romannumeral2}  & 0.6496  & 0.6594\\
		QMWF-LM  & 0.7520 & 0.8140 &QMWF-LM  & 0.6950 & \textbf{0.7100}\\
		CNM  & 0.7701 & \textbf{0.8591} &CNM  & 0.6748 & 0.6864\\
		QLM-EE  & \textbf{0.7713} & 0.8542 &QLM-EE  & \textbf{0.6956} & 0.7003\\
		\toprule[1pt]
	\end{tabular}
\end{table}
\subsubsection{Hyper-parameters}
Cosine similarity defined by (\ref{cosreal}) is used as the distance metric between the real-valued feature vectors after pooling. The hinge loss \cite{hu2014convolutional} for training the model is given by
\begin{equation}
\mathcal L= \rm{max}\{0, 0.1-\mathcal C_{+} + \mathcal C_{-}\},
\end{equation}
where $\mathcal C_{+}$ is the cosine similarity of a ground truth answer, $\mathcal C_{-}$ is the cosine similarity of an incorrect answer randomly chosen from the entire answer space.

\begin{table*}[!t]
	\renewcommand{\arraystretch}{1.2}
	\caption{Ablation results on TREC-QA. QLM with separable-state embedding (QLM-SE) and QLM with mixed-state embedding (QLM-ME) are compared with QLM-EE. QLM-EE-Real is the real-valued QLM-EE.}
	\label{Ablation}
	\centering
	\begin{tabular}{c|cccccc}
		\toprule[1pt]
		\textbf{Model} & $\bm D$& $\bm N$ & \textbf{MAP} & \textbf{MRR} & \textbf{Params.} & \textbf{FLOPs} \\
		\hline 			
		{QLM-EE} 	&	8  & 2 & \textbf{0.7302$\pm$0.0101} & 0.8071$\pm$0.0156 & 1.45M & 4.20M\\
		\hline
		{QLM-ME} 	&	8  & 2 & 0.6905$\pm$0.0711 & 0.7584$\pm$0.0059 & 0.99M & 4.81M\\
		\hline
		{QLM-ME} 	&	64  & 2 & 0.7267$\pm$0.0194 & \textbf{0.8171$\pm$0.0151} & 7.88M & 124.56M\\
		\hline
		{QLM-EE-Real} 	& 16  & 2 & 0.6778$\pm$0.0139 &0.7505$\pm$0.0187 & 1.97M & 6.01M\\
		\hline
		{QLM-SE} 	&	8  & 2&0.6934$\pm$0.0149 & 0.7804$\pm$0.0156   & 1.32M & 3.28M\\
		\toprule[1pt]
	\end{tabular}
\end{table*}

The parameters in the QLM-EE are determined by the set of hyper-parameters $\Theta:=\{ N, D, M \}$, where $N$ is the number of words in a sequence, $D$ is the word embedding dimension and $M$ is the number of measurement vectors. $N=1$ means embedding the words without composition, and in this case no entanglement can be generated. We test single-layer and two-layer fully connected neural networks with $\{128, 256, 512\}$ neurons for entanglement embedding. A grid search is conducted using $N \in \{1, 2, 3\}$, $D \in \{6, 8, 10, 12, 14, 16\}$, $M \in \{500, 1000, 2500, 5000 \}$, batch size in $\{16, 32, 64\}$ and learning rate in $\{0.01, 0.1, 0.5\}$. In line with CNM, the concatenation of the feature vectors for $N=1,2,3$ has also been tested. Larger $N$ has been tried, but $N \in \{1, 2, 3\}$ shows better performance than $N\geq4$. All the parameters are initialized from standard normal distributions except the measurement vectors, which are initialized by orthogonal vectors.

\subsection{Performance}
The experimental results on TREC-QA and $\rm W_{IKI}$QA are shown in TABLE~\ref{Result}. We compare the performances of the classical models, including CNNs, Recurrent NNs and attention models with the performances of QLMs. QLM-EE is consistently better than all the QLMs and classical models on both datasets if MAP is used as the metric. If we use MRR as the metric, QLM-EE performs slightly worse than CNM on TREC-QA while better than the other models, and slightly worse than QMWF-LM while better than all the other models on $\rm W_{IKI}$QA.

Our word embedding dimension is selected from $D \in \{6, 8, 10, 12, 14, 16\}$, which is much smaller than the word embedding dimension of previous QLMs selected from $\{50,100,200\}$. As a consequence, the amount of parameters for the word embedding layer has seen dramatic reduction while the performance of model has been improved. Fig.~\ref{fig_sim} illustrates how the word embedding dimension affects the performances of different models in terms of MAP and MRR on TREC-QA. It is clear that the model with the concatenation of the feature vectors for $N = 1,2,3$ performs best and the best word embedding dimension is $8$, which is significantly smaller than $\{50,100,200\}$ for the previous QLMs.

As $D$ increases, the dimension of the state vector after the entanglement embedding is increased according to the formula $D^N$. As pointed out in the literature \cite{yin2018dimensionality}, learning word embedding has the risk of underfitting or overfitting. Embedding dimension that is too small (less than 50) or too large (more than 300) will degrade the performance. In QLM-EE, the $N$-gram lies in the tensor-product space, which tends to result in a high-dimensional representation and poor performance. For example, if the word embedding dimension is 8, then $3$-grams are described by complex-valued $512$-dimensional vectors, or real-valued $1024$-dimensional vectors. In addition, the dimension of EE module and measurement vectors will increase accordingly, leading to a model that easily overfits the data. In our case, the performance of $3$-gram model is worse than that of the $2$-gram model, which is a sign of overfitting in this relatively small-sized dataset. However, it is possible that high-dimensional representation will improve the performance if a much larger dataset is considered.

\begin{figure*}[!t]
	\centering
	\includegraphics[width=7in]{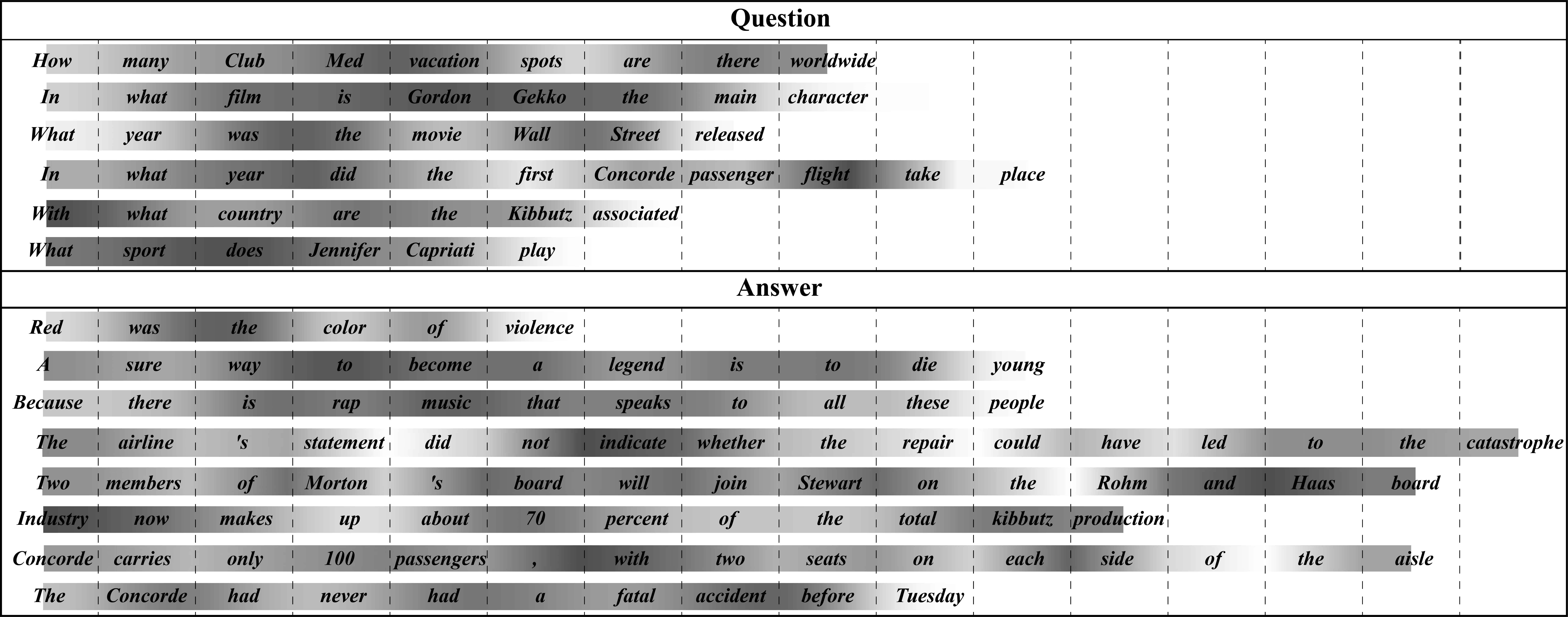}
	\caption{Entanglement entropy of the 2-grams in the sentences. Darker color is an indication of larger entanglement between adjacent words. }
	\label{Ent_deg_words}
\end{figure*}

\subsection{Ablation Test}

Ablation test studies the contribution of certain components to the overall performance by removing these components from the model. We conduct ablation tests to evaluate the effectiveness of entanglement embedding on 2-gram QLM.  QLM-SE has removed thxe entanglement embedding module and measurements are directly performed on the separable joint states. QLM-ME generates the mixed-state embedding by (\ref{ms}), and measurements are performed on the density matrix $\rho$. In the setting of complex-valued QLM-EE, we let $D=8$ and $M=3000$. For the complex-valued QLM-ME whose structure is the same as CNM, we set $D=8$ and $D=64$ with $M=3000$ for comparison. Note that when $D=64$, the complex-valued QLM-ME has the same dimension for $2$-grams as QLM-EE. However, since the word embedding dimension has been increased, the number of parameters for the word embedding module has been increased accordingly. We have doubled the word embedding dimension to $D=16$ in the real-valued QLM-EE (QLM-EE-Real), and thus the dimension of $2$-grams is $256$ which is four times larger than the dimension of $2$-grams in the complex-valued QLM-EE. Besides, the dimension of the measurement vectors has been increased four times accordingly. To make the number of parameters compatible, we set $M=1500$ for this case.

\begin{table}[!t]
	\centering
	\renewcommand{\arraystretch}{1.2}
	\caption{The two-sample K-S test statistics $k$ at $p$ significance level between the performance metrics of QLM-EE ($D=8$) and other models in ablation test. $k = 1$ indicates that the two sets of metric data are sampled from different distributions.}
	\label{K-Stest}
	\begin{tabular}{c|c|c|c|c}
		\toprule[1pt]
		\textbf{Model} &$k_{MAP}$ & $p_{MAP}$ &$k_{MRR}$&$p_{MRR}$ \\
		\hline
		\textbf{QLM-ME ($D=8$)}  & 1& $0.17\%$& 1& $0.02\%$ \\
		\hline
		\textbf{QLM-ME ($D=64$)}  & 1& $3.18\%$ & 1& $6.75\%$ \\
		\hline
		\textbf{QLM-SE ($D=8$) }  & 1 & $0.01\%$& 1& $0.02\%$\\
		\hline
	    \textbf{QLM-EE-Real ($D=8$)}  & 1& $1.12\%$& 1& $1.18\%$\\	
		\hline
		\toprule[1pt]
	\end{tabular}

\end{table}

We have run the experiment 10 times. The mean and standard deviation of metrics, the number of parameters and Floating Point Operations (FLOPs) are reported in TABLE~\ref{Ablation}. We can see that QLM-EE achieves the best performance for 2-gram model when the word embedding dimension is $8$. The performances QLM-ME with $D=64$ and QLM-EE with $D=8$ are close, but the number of parameters and FLOPs used by the former is significantly greater than the latter. The result of QLM-EE-Real confirms that the complex-valued neural network indeed promotes the performance of QLMs. The two-sample Kolmogorov-Smirnov (K-S) test result on the performance metrics is shown in TABLE~\ref{K-Stest}, which verifies the performance differences of these models.


\section{Post-hoc Interpretability}
von Neumann entanglement entropy $\mathcal S$ \cite{horodecki1994quantum} is an accurate measure of the degree of quantum entanglement for a bipartite quantum pure state. The entanglement entropy is calculated as follows
\begin{equation}
{\mathcal S} = -\sum^{K}_{i = 1}  \arrowvert\lambda_i\arrowvert^2 {\rm log} ( \arrowvert\lambda_i\arrowvert^2),
\label{Von_Neumann_entropy}
\end{equation}
where $\lambda_i$ is the Schmidt coefficient of the composite pure state and $K$ is the minimal dimension of the subsystems, i.e., $K={\rm min}({\rm dim}(\mathcal{H}_A),{\rm dim}(\mathcal{H}_B))$. Apart from the analytical entanglement measure (\ref{Von_Neumann_entropy}) for bipartite states, it is also worth mentioning that efficient numerical methods are available for quantifying quantum entanglement for multipartite states \cite{zhang2020iterative}.

\begin{table}
	\renewcommand{\arraystretch}{1.2}
	\caption{Selected entangled 2-grams in TREC-QA}
	\label{S_ent_words}
	\centering
	\begin{tabular}{p{1.8cm}|p{5.5cm}}
		\toprule[1pt]
		\textbf{Type} & \textbf{Word combinations} \\
		\hline
		Most entangled in questions  & annual~revenue; as~a; is~cataracts; how~long; tale~of; first~movie; ethnic~background\\
		\hline
		Least entangled in questions  & introduced~Jar; the~name; what~year; the~main; who~is; whom~were ; is~a; in~what\\
		\hline
		Most entangled in answers  & ends~up; never~met; plane~assigned; in~kindergarten; academy~of; going~to; agricultural~farming; secure~defendants \\
		\hline
		Least entangled in answers & skinks~LRB; while~some; grounded~in; of~Quarry; of~seven; he~said; in~China; responsibility~and \\
		\toprule[1pt]
	\end{tabular}
\end{table}

\begin{table}
	\renewcommand{\arraystretch}{1.2}
	\caption{Selected entangled 3-grams in TREC-QA}
	\label{S_ent_words1}
	\centering
	\begin{tabular}{p{1.8cm}|p{5.5cm}}
		\toprule[1pt]
		\textbf{Type} & \textbf{Word combinations} \\
		\hline
		Most entangled in questions  & the company Rohm; Hale Bopp comet; how often does; Insane Clown Posse; Capriati~'s~coach\\
		\hline
		Least entangled in questions  & there worldwide ?; What is Crips; Criminal Court try; When did James\\
		\hline
		Most entangled in answers  & after~a~song; comet~'s~nucleus; Out~of~obligation; at~times .; Black~Panther~Party \\
		\hline
		Least entangled in answers & Queen , Tirado; came from Britain; Nobel Prize last; appears in the; Americans over 50\\
		\toprule[1pt]
	\end{tabular}
\end{table}

TABLE~\ref{S_ent_words} shows the selected most and least 2-grams of words, ranked by the von Neumann entanglement entropy. The most entangled pairs are mostly set phrase or some well-known combinations of words, e.g., \textit{how~long}. However, \textit{is cataracts} is clearly not a set phrase or well-known combination of words. We found that \textit{is cataracts} appears many times, and \textit{cataracts} is always next to \textit{is} in this particular training dataset. This may be the reason why the learned model takes \textit{is cataracts} as a fixed combination of words. The least entangled pairs consist of words with fixed semantic meaning such as names $\{$\textit{Quarry}, \textit{China}$\}$ and interrogatives $\{$\textit{what}, \textit{who}$\}$, some of which appear only once in the dataset. In other words, there is not so much semantic ambiguity or superposition with these combinations that demands a quantum probabilistic interpretation.

TABLE~\ref{S_ent_words1} shows the selected most and least entangled 3-grams, whose von Neumann entanglement entropies are calculated to indicate the entanglement between the first two words, and the remaining word (the third word). Similar to 2-grams, the most entangled 3-grams are fixed collocations and combinations that often appear together in the training set, such as a set phrase. Interestingly, punctuation marks also appear in the most entangled 3-grams. For example, entanglement in \textit{at~times .} is large, which implies that the phrase \textit{at~times} is often at the end of the answers. However, \textit{there worldwide ?} is among the least entangled combinations, since \textit{there worldwide} is not in any of the question sentences for training. The third word in the least entangled 3-grams are mainly names and numbers, which cannot combine with the first two words to form a fixed phrase.

In Fig.~\ref{Ent_deg_words}, we also visualize the entanglement entropy in some selected sentences, where the degree of darkness indicates the level of entanglement. It can be seen that words with multiple meanings in different contexts, e.g., $\{$\textit{is}, \textit{does}, \textit{the}, \textit{film}$\}$, have greater capabilities to entangle with their neighboring words.


\section{Conclusion}
In this paper, we proposed an interpretable quantum-inspired language model with a novel EE module in the neural network architecture. The EE enables the modelling of the word sequence by a general quantum pure state, which is capable of capturing all the classical and non-classical correlations between the word states. The expressivity of the neural network is greatly enhanced by cascading the word embedding and EE modules. The complex-valued model has demonstrated superior performance on the QA datasets, with much smaller word embedding dimensions compared to previous QLMs. In addition, the non-classical correlations between the word states can be quantified and visualized by appropriate entanglement measures, which improves the post-hoc interpretability of the learned model.

The future plan is to apply the adaptive methods \cite{huszar2012adaptive,qi2017adaptive} for optimizing the virtual measurement operations to increase the efficiency in feature extraction. The QLM-EE model is expected to be more powerful on huge datasets in which the semantic meanings of words and their correlations are far more complex. With a larger dataset and richer semantic superpositions between the words, several entanglement embedding modules can be cascaded to form a deeper neural network, which could encode the multipartite correlations within the text at different scales.

%
%
%


\ifCLASSOPTIONcaptionsoff
  \newpage
\fi

\bibliographystyle{IEEEtran}
\bibliography{ref}

\end{document}